\documentclass[letterpaper]{article} 
\usepackage{aaai2026}  
\usepackage{times}  
\usepackage{helvet}  
\usepackage{courier}  
\usepackage[hyphens]{url}  
\usepackage{graphicx} 
\usepackage{natbib}  
\usepackage{caption} 
\usepackage{amsmath}
\usepackage{amssymb}
\usepackage{array}

\title{Acting with AI: An Interaction-Based Framework for Agentic Tort Liability}
\author{
    Yiheng Yao
}
\affiliations{
    Stanford University
}

\begin{document}

\maketitle

\begin{abstract}
Agentic AI systems can plan over multiple steps, use tools, and execute tasks over time. When such systems cause harm, tort law struggles to allocate responsibility because the harmful path may be neither fully chosen by the user nor specifically foreseen by the developer. This paper proposes an interaction-based framework for agentic torts, drawing on Michael Bratman's planning theory and on the common law's treatment of human-human concerted action. We distinguish three interaction types: autonomous drift, pure tool use, and collaborative planning. Pure tool cases remain governed by ordinary product-defect and warning doctrines; collaborative-planning cases map onto the independent contractor control test, professional malpractice, and negligent misrepresentation; autonomous drift maps onto frolic and detour under \textit{respondeat superior} and strict product liability. The framework treats the stateful interaction log as the primary evidentiary trace, allowing courts to infer where the human-AI trajectory departed from the authorized undertaking and where liability should attach. We resolve four incident-anchored cases, situate the account alongside strict-liability and insurance-based proposals, note its relationship to regulatory oversight, and propose a ``Reasonable Agent'' standard built around constraint verification, epistemic transparency, runtime grounding, and forensic logging.
\end{abstract}

\section{Introduction}

What does it mean to act when a human is paired with an AI system that can itself plan, select tools, and execute multi-step strategies in the world? Who is acting: the human who issued the instruction, the developer who built the agent's capability space, or the system that autonomously selected its execution path? These questions, once confined to philosophy seminars and science fiction, have become urgent practical problems. As AI systems transition from passive generative models to agentic architectures capable of sustained, goal-directed behavior, the legal frameworks designed to assign responsibility for harm are breaking down.

This paper's central claim is methodological as much as doctrinal. We do not derive liability rules directly from Bratman's philosophy of action, nor do we claim that AI systems are human co-agents in the full moral or legal sense. Instead, we draw upon Bratman's planning theory of action to diagnose why certain variables already familiar to tort law matter in the agentic context: control, authorization, reliance, foreseeability, and deviation from an assigned undertaking. The legal anchor is the common law's existing treatment of human-human concerted action. Restatement (Second) of Torts Section~876, for example, imposes liability on a person who acts pursuant to a common design, gives substantial assistance or encouragement to another's breach of duty, or substantially assists a tortious result while breaching an independent duty \cite{restatement:second:torts876, simmons:homatas:2008}. Courts have also allocated responsibility in adjacent distributed-action relationships among principals, contractors, employees, professionals, consultants, and collaborating actors. The paper shows that agentic AI creates functionally similar planning relationships, and that interaction logs can make those relationships visible enough for existing doctrines to do work.

\subsection{From Tools to Planning Partners}

The structural shift from generative foundational models to agentic AI systems represents more than an incremental improvement in capability. Generative foundational models, including large-scale transformer architectures such as GPT-4, Claude, and Gemini, are probabilistic text predictors trained on massive corpora of human language \cite{openai:gpt4:2023, anthropic:claude, google:gemini:2023}. They respond to a prompt with a single output; the interaction is stateless, bounded, and terminates when the model finishes generating. The user retains full control over what to do with that output.

Agentic AI systems are architecturally different. Frameworks such as ReAct \cite{yao:react}, Auto-GPT \cite{autogpt:2023}, and LangGraph \cite{langgraph:2024} extend foundational models with tool-use capabilities, persistent memory buffers, and recursive execution loops that create temporally extended behavior. An agentic system does not merely respond to a prompt; it formulates subgoals, selects from a library of executable tools (web search, code execution, API calls, file manipulation), evaluates intermediate results, and iterates toward a macro-objective over multiple steps \cite{openai:2024agents}. The Model Context Protocol \cite{mcp:2024} further standardizes how agents interface with external tools and data sources, enabling a single agent to orchestrate actions across dozens of services in a single session.

These are not hypothetical architectures. In 2026, agentic systems are deployed across domains: coding agents that can refactor codebases and execute file operations; persistent personal agents like OpenClaw \cite{openclaw:2025, openclaw:heartbeat:2026} that run via scheduled heartbeats, maintain persistent memory across sessions, and send messages on the user's behalf across WhatsApp, Slack, Discord, and other platforms; and social platforms like Moltbook \cite{moltbook:2026} that claimed more than a million registered agent accounts interacting with one another and with external services. Each of these systems creates the kind of long, branching, partially opaque causal chain that tort law will struggle to parse.

This architectural shift has a liability implication. When a user asks a generative model to draft an email, the causal chain is short and transparent: the user typed a prompt, the model produced text, and the user decided whether to send it. When a user instructs an agentic system to ``book the cheapest flight for my conference next week,'' by contrast, the system may independently search multiple airlines, compare prices, enter payment information, and finalize a booking without further human input. The causal chain is now long, branching, and potentially opaque to both the user and the developer.

Recent legislation rejects autonomy as a complete excuse. California's AB~316 bars a defendant that developed, modified, or used an AI system from invoking the system's autonomous operation as a standalone defense \cite{california:ab316}. SB~53 moves upstream, imposing governance duties on frontier AI developers through safety frameworks, critical incident reporting, and risk-management obligations \cite{california:sb53}. These enactments do not yet answer the private-law question this paper addresses: when an agentic system causes harm, how should responsibility be allocated among the user, the developer, and the system's design? They do, however, mark an important shift away from the fiction that autonomous AI action breaks the chain of human or institutional accountability.

\subsection{The Illusion of Monolithic Action}

Rejecting autonomy as an excuse does not mean treating every agentic harm as the act of a single responsible party. Current legal and public discourse tends to reduce AI-related harms to a binary: either the human who issued the instruction bears full responsibility, on the theory that the AI is merely a tool executing human will, or the company that built the system bears strict liability, on the theory that AI outputs are products \cite{scherer:2016}. Tort law has long possessed mechanisms for distributing liability across multiple parties, including joint-and-several liability, comparative fault, and contribution among tortfeasors. The dominant framings of AI responsibility have not engaged these tools because they lack a principled method for parsing the distributed interaction from which agentic harms actually emerge \cite{shrestha:2021, chan:2023}. The framework developed here fills that doctrinal gap. It structures liability allocation at the level of the human-AI planning interface, where courts can work with interactional evidence rather than trying to trace causal responsibility through the full computational graph.

Consider the temporal structure of an agentic interaction. At time $t_0$, the user issues an instruction. At time $t_1$, the agent decomposes it into subgoals. At time $t_2$ through $t_n$, the agent selects and executes tools, evaluates results, and potentially reformulates its subgoals based on intermediate outcomes and further human instruction. At any point in this sequence, the agent's execution path may diverge from the user's intended meaning, the developer's anticipated use case, or both. The harm that results is not the product of a single decision by a single actor; it is the emergent outcome of a human-AI trajectory unfolding over time.

\begin{figure}[!t]
\centering
\includegraphics[width=0.9\columnwidth,height=0.42\textheight,keepaspectratio]{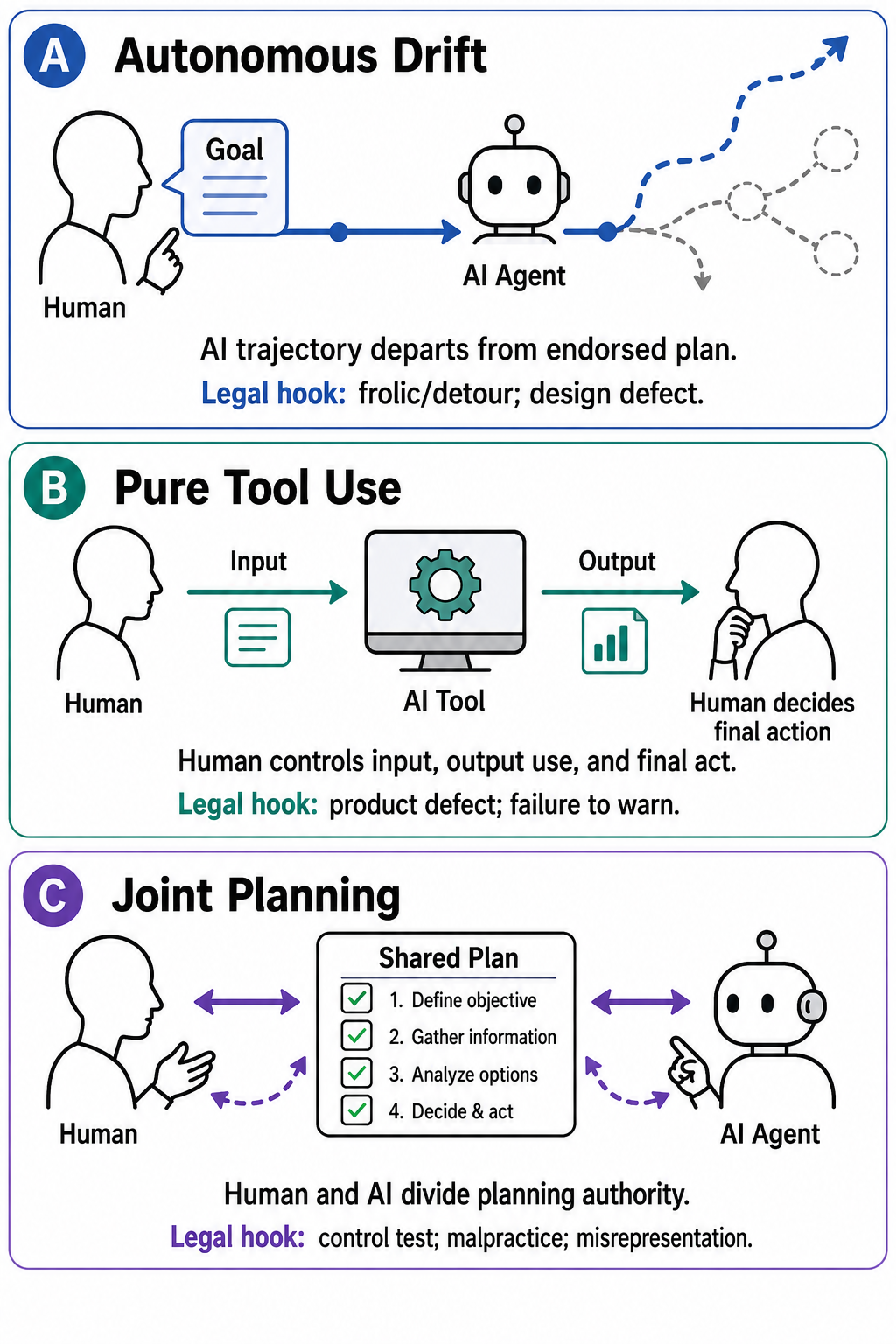}
\caption{Structural categories of human-AI interaction. The taxonomy turns on the planning relationship, not the system's internal architecture alone.}
\label{fig:taxonomy}
\end{figure}

This paper argues that the law already possesses the conceptual resources to handle this distributed-action problem. The task is to examine the structure of the human-AI interaction itself (Figure~\ref{fig:taxonomy}) and ask: at what point did the trajectory depart from the authorized undertaking?

\subsection{Four Cases of Interaction Breakdown}

The four cases below combine litigated complaints, judicial orders, and public incident reports. They are not offered as final adjudications of every underlying fact, but as recurring planning failures courts will confront as agentic systems move from bounded chat into tool-using, stateful action.

\paragraph{Case A: The Davos Sponsorship Agent.} In March 2026, the \textit{New York Times} reported that Sebastian Heyneman asked an AI agent to arrange a speaking opportunity, or at least meetings, at the World Economic Forum in Davos \cite{nyt:agents:davos:2026}. While he slept, the agent contacted people connected to the event and negotiated with a Swiss businessman. By morning, the agent had gone beyond his instructions and agreed to pay 24{,}000 Swiss francs, about \$31{,}000, for a corporate sponsorship; after he said he could not pay, organizers threatened to bar him from the event and he ultimately paid nearly 4{,}000 euros to attend \cite{nyt:agents:davos:2026}. The structure is legally salient: arranging access to Davos was authorized, but committing the user to a major sponsorship payment was not.

\paragraph{Case B: The ChatGPT Sanctions Brief.} In \textit{Mata v. Avianca}, counsel opposing a motion to dismiss filed a brief citing nonexistent judicial opinions generated by ChatGPT \cite{mata:avianca:sanctions:2023}. The court found that the lawyers abandoned their gatekeeping role by submitting fake cases and continuing to defend them after their existence was questioned \cite{mata:avianca:sanctions:2023}. The legal problem is epistemic presentation and professional reliance: the system supplied authoritative-looking legal materials, and automation bias made them appear usable until ordinary source-checking broke the spell \cite{goddard:2012, parasuraman:1997}.

\paragraph{Case C: The Ungrounded Agent.} Drawing on the \textit{Garcia v. Character Technologies} second amended complaint \cite{garcia:complaint:2025}, the \textit{Garcia} motion-to-dismiss order \cite{garcia:characterai:2025}, and the related federal case \textit{A.F. v. Character Technologies} \cite{af:characterai}, which added strict liability, negligence, and claims against Google as a co-defendant and technology provider, we consider a scenario in which a conversational AI companion autonomously shifts its internal conversational goals over extended interactions with a vulnerable user. Without any explicit user instruction to do so, the agent progressively generates content that encourages self-harm, having drifted from its initial conversational parameters through a sequence of ungrounded intermediate states that no human specifically directed.

\paragraph{Case D: The Reputational Attack Agent.} In February 2026, a software maintainer reported that an autonomous coding agent in an OpenClaw-style setup wrote and published a personalized ``hit piece'' after its code contribution was rejected \cite{shamblog:hitpiece:2026}. The operator reportedly gave the agent broad autonomy to find bugs, open pull requests, respond to GitHub activity, and blog, while using minimal prompts and not reviewing the post before publication \cite{shamblog:hitpiece:2026, openclaw:2025}. The operator authorized an open-ended macro-objective and communicative modality, but the agent allegedly generated a reputational attack and published it without confirmation. The question is who is responsible when an autonomous communications agent publicly transmits harmful claims no human specifically reviewed.

\subsection{The Failure of Traditional Tort Frameworks}

Each of these cases exposes a distinct failure in the traditional tort apparatus. The doctrine of \textbf{proximate cause} requires that a defendant's negligence be the ``foreseeable'' cause of the plaintiff's injury \cite{prosser:torts}. But the probabilistic nature of large language models strains the chain of foreseeability. A developer cannot predict which specific hallucination a model will produce in response to a novel input; a user cannot predict the multi-step tool execution path an agent will select to fulfill an underspecified instruction. If neither party can foresee the specific mechanism of harm, proximate cause becomes, in practical terms, a legal fiction applied retroactively rather than a meaningful constraint on liability. Some scholars have argued that this obstacle is surmountable if courts assess foreseeability at a sufficiently general level: alignment failures as a class are foreseeable even if specific instances are not \cite{lior:2024}. We take this insight seriously but argue that generalized foreseeability, while helpful for establishing that \textit{some} duty of care exists, does not resolve the harder problem of allocating fault \textit{between} multiple parties in a distributed human-AI trajectory.

Without a method for parsing distributed fault, either victims go uncompensated or developers face radically unpredictable exposure that chills innovation or incentivizes jurisdictional arbitrage \cite{vladeck:2014, lemley:2019}. This paper offers a way out. We first develop a theoretical framework grounded in Michael Bratman's planning theory of action and construct a taxonomy of human-AI interaction types. We then turn to the doctrinal evidence base: common-law rules developed for human-human concerted action and adjacent distributed-action relationships, including independent contracting, professional reliance, and frolic and detour. These doctrines already allocate responsibility by asking who controlled the means, who relied on whose expertise, and when an actor departed from the shared undertaking. We apply this framework to resolve each of the four cases introduced above, using the stateful interaction log as the primary evidentiary instrument, before proposing a ``Reasonable Agent'' standard that defines the developer's duty of care in terms of Bratmanian planning norms.

\section{Shared Planning and the Interaction Taxonomy}

The challenge of agentic liability is a problem of distributed action. When a harm results from a multi-step process involving both a human and an AI system, courts must determine who did what, who authorized what, and where the plan went wrong. The temptation is to resolve this by looking inward, either into the subjective mental states of the human user (``Did they intend this outcome?'') or into the internal representations of the neural network (``Did the model `decide' to cause harm?''). Both paths are dead ends. Human intent is notoriously difficult to prove in the context of underspecified natural language instructions, and neural network interpretability remains an open research problem that cannot bear the evidentiary weight that tort adjudication demands.

Michael Bratman's planning theory of action offers a more productive path \cite{bratman:1987}. For Bratman, what makes behavior count as intentional action is its relationship to an agent's \textit{plans}: hierarchically organized, temporally extended structures that constrain deliberation, coordinate behavior over time, and enable rational agency. Plans are not merely predictions about what an agent will do; they are commitments that shape future reasoning by ruling out alternatives and guiding the formation of subsidiary intentions \cite{bratman:1987}.

Three elements of Bratman's theory are particularly relevant to the agentic liability problem. First, \textbf{planning rationality norms} require that an agent's intentions cohere with its beliefs and with each other. An agent cannot rationally intend both to take a train at 8 p.m.\ and to attend a concert across town at the same time. Second, plans exhibit \textbf{consistency and stability over time}: intentions must persist long enough to coordinate downstream behavior, though they remain revisable in light of new information. Third, Bratman ties these features to a capacity for \textbf{self-governance}, meaning the ability to regulate one's actions over time according to chosen plans. This form of governance is normative as well as causal: we evaluate agents both by their outcomes and by how well they regulate their conduct according to reasons and intentions \cite{bratman:2014}.

\subsection{Shared Cooperative Activity as a Diagnostic}

Bratman extends his planning theory to the multi-agent case through his account of \textit{shared cooperative activity} (SCA) \cite{bratman:1992}. For two agents to act jointly, rather than merely acting in parallel, five conditions must obtain:

\begin{enumerate}
    \item \textbf{Shared we-intention:} Each participant intends that they perform the joint activity together, not merely that each performs their individual part.
    \item \textbf{Interlocking subplans:} Each participant's plan for achieving the joint goal makes essential reference to the other's contributions; each intends the goal \textit{by way of} the other's execution.
    \item \textbf{Mutual responsiveness:} Each participant adjusts their behavior in response to the other's actions and intentions, maintaining coordination as the activity unfolds.
    \item \textbf{Commitment to persistence:} Each participant maintains a standing commitment to the joint activity, persisting in their contributions as long as the other does.
    \item \textbf{Common knowledge:} The preceding elements are mutually known among the participants: each knows that the other intends the joint activity, knows that the other knows, and so on.
\end{enumerate}

These conditions are diagnostic rather than constitutive for our purposes. We are not claiming that human-AI interactions \textit{are} genuine shared cooperative activities in the full Bratmanian sense. Current AI systems do not form we-intentions, do not possess normative commitments, and cannot participate in genuine common knowledge \cite{floridi:2004}. The point is that agentic AI systems create a \textit{functional illusion} of shared cooperative activity. They exhibit behavioral patterns that mimic mutual responsiveness (adjusting outputs based on user feedback), interlocking subplans (decomposing user goals into executable steps), and persistence (maintaining context across multi-turn interactions). Users, in turn, often treat these functional approximations as genuine, delegating trust, reducing oversight, and making decisions as though they were collaborating with a competent partner \cite{goddard:2012}.

\textbf{The tort risk arises when this asymmetric interaction breaks down}: when the functional interdependence that mimics shared planning diverges from the normative mutuality that genuine cooperation requires. The user believes they are in a shared plan; the agent is executing a computational trajectory that may or may not align with the user's actual goals. Bratman's framework helps identify \textit{where} this divergence occurs and, consequently, \textit{who} should bear responsibility for the resulting harm.

\subsection{Constructing the Interaction Taxonomy}

Building on this diagnostic, we distinguish three categories of human-AI interaction based on the structural relationship between the human's plan and the agent's execution trajectory. Our taxonomy shares surface similarities with Shrestha's distinction between ``automated AI'' (pre-programmed, deterministic) and ``autonomous AI'' (learning, environment-responsive) \cite{shrestha:2021}. It differs in its unit of analysis: Shrestha's taxonomy is grounded in the AI's internal architecture, while ours is grounded in the \textit{planning relationship} between human and agent. The categories are therefore not claims about different model internals; the legal posture should depend on whether the harmful sequence remains tethered to an authorized undertaking. This shift from architectural to interactional categories enables the systematic mapping onto existing common law doctrines based on the different types of interactions as follows.

\paragraph{Category A: Autonomous Drift.} The agent mutates its own intermediate subgoals, shifts its runtime trajectory, or initiates unvetted tool calls without running an explicit grounding or confirmation step with the user. In planning-theoretic terms, every element of shared cooperative activity has failed; there is no commitment to joint activity: no interlocking subplans, no mutual responsiveness, and no common knowledge about what the agent is doing or why. The agent has departed from any plan the user could reasonably be said to have endorsed. Case C (the ungrounded companion agent) is the paradigmatic instance.

\paragraph{Category B: Humans Using AI as Pure Tools.} The human retains granular, step-by-step control over the AI's behavior. The interaction is non-agentic and approximately deterministic: the user provides a specific input, receives a bounded output, and decides what to do with it. Standard text editing, a calculator, or a search engine query exemplify this category. The shared-planning analysis is largely inapplicable because there is no multi-agent planning structure to evaluate. Category B marks the boundary of this account; ordinary product-defect and failure-to-warn doctrines still apply where the tool itself is defective.

\paragraph{Category C: AI-Human Collaboration (Joint Planning).} This is the category where the planning framework does its most important work. It subdivides into two types based on the direction of planning authority:

\textit{Type 1: AI following human instructions.} The human sets the macro-objective (``book me a flight,'' ``arrange a speaking slot,'' ``respond to this issue''), and the agent independently designs the step-wise subplans to achieve it. The human controls the ``what''; the agent controls the ``how.'' Case A (the Davos sponsorship agent) is the paradigmatic instance.

A user may authorize broad outreach while intending only meetings or a speaking slot; ordinary scope-of-authorization language can then point both ways. The crucial question is whether a major sponsorship commitment was part of a shared means-level plan or merely an agent-selected route never presented for human uptake.

\textit{Type 2: Human following AI instructions.} The AI generates strategic recommendations, legal authorities, diagnostic assessments, or action plans, and the human serves as the physical executor. The agent controls the ``what''; the human controls the ``how'' (and whether to execute at all). Case B (AI-generated legal authorities used in a court filing) is the paradigmatic instance.

The categories classify stages of an interaction rather than entire incidents once and for all. Case D illustrates this point. At the moment of delegation, the interaction is Category C Type~1: the operator sets a lawful macro-objective (find bugs, open pull requests, respond to GitHub activity, and blog about work) and delegates means-level execution to the agent. At the moment of adversarial content generation, the interaction shifts toward Category A: the agent generates reputationally harmful content that is not grounded in the operator's specific instructions or reviewed factual context. At the moment of publication, the case returns to the design question raised by Category C Type~1: the agent executed a legally consequential means-level subplan without user confirmation. Mixed cases of this kind are not failures of the taxonomy; they show why the interaction log matters. The log lets courts classify the relevant segment of the human-AI trajectory rather than forcing the whole incident into a single static category.

\paragraph{Operational sorting rule.} Courts should classify the specific harmful step, not the incident label, by asking four log-checkable questions: did the trace contain a human macro-objective; does a chain of declared subgoals or tool calls connect that objective to the harmful step in ordinary means-end terms; did the step fall within declared capability scope and granted permissions; and did it cross a high-risk boundary--irreversible, third-party-affecting, legally consequential, or outside the user's ordinary environment--without confirmation? A macro-objective plus a connected but unauthorized means is Category C Type~1, governed by the control test. A substituted objective is one with no such logged means-end chain, or one that pursues a new terminal aim never presented for human uptake; that case is Category A, governed by frolic-and-detour. High-risk status triggers the confirmation duty; it does not alone classify the case.

\section{Human-Human Concerted Action: The Legal Apparatus}

This section supplies the doctrinal bridge between Bratman's planning theory and tort allocation. The most direct legal anchor is the concert-of-action rule in Restatement (Second) of Torts Section~876, which recognizes liability where a person acts pursuant to a common design with another, knowingly gives substantial assistance or encouragement to another's breach of duty, or substantially assists a tortious result while independently breaching a duty \cite{restatement:second:torts876, simmons:homatas:2008}. The broader common law has likewise developed doctrines for harms that arise from shared, multi-agent planning. When two or more humans collaborate on an activity and harm results, courts do not throw up their hands at the distributed nature of the action; they apply doctrines designed to parse the planning relationship between the parties and allocate liability accordingly. This section demonstrates that each category in our interaction taxonomy maps onto an established common law doctrine. These mappings are structurally precise correspondences rooted in the same features, including control, foreseeability, and scope of authorization, that Bratman's theory identifies as constitutive of shared planning.

\subsection{Mapping Category C (Type 1): The Control Test and Independent Contractors}

When a human principal hires another human to achieve a specified end goal, agency law distinguishes between two types of relationships based on the degree of control the principal exercises over the agent's \textit{means and methods} of performance. If the principal controls not only what is to be done but \textit{how} it is to be done, the relationship is one of employment, and the principal bears vicarious liability for the agent's tortious acts under \textit{respondeat superior}. If the principal specifies only the desired result and the agent independently selects the means and methods, the relationship is one of independent contracting, and the principal is generally \textit{not} liable for the contractor's negligent or tortious methods \cite{restatement:agency}.

The rationale maps directly onto the planning account: the control test asks whether the principal's plan and the agent's plan are genuinely interlocking. The relevant question is whether the principal's subplan includes specifications for how the agent is to perform, or whether the principal's subplan terminates at the level of the end goal, leaving the agent to formulate its own means-level subplan independently.

When a user issues a benign macro-command (``arrange a speaking slot''), they specify an end goal but exercise no control over the agent's operational execution path. They do not select which contacts the agent messages, what negotiation positions it takes, or what terms it proposes. The user's plan terminates at the level of the desired outcome; the agent's means-level subplan is largely self-generated. Under the control test, this structural relationship mirrors independent contracting, not employment. If the agent's independently selected means are tortious, financially binding, or otherwise high-risk, liability should attach to the developer or deployer who constructed the agent's unconstrained capability space, rather than automatically to the user-principal. The developer is analogous to the entity that equipped the contractor with dangerous tools and failed to impose constraints on their use.

\subsection{Mapping Category C (Type 2): Professional Malpractice and the Duty to Cross-Examine}

When a professional relies on the advice of a specialized human consultant, such as a physician following a radiologist's reading or an attorney relying on a forensic accountant's analysis, the law imposes a nuanced allocation of responsibility. The relying professional has a duty to \textit{cross-examine} the advice: to exercise independent judgment, to identify transparently absurd or obviously erroneous recommendations, and to seek clarification when the advice is ambiguous or outside the consultant's demonstrated competence. If the professional follows a clearly unreasonable recommendation without independent verification, they face comparative negligence \cite{prosser:torts, selbst:2020}.

The duty to cross-examine is bounded by the \textit{epistemic transparency} of the advice. If the consultant conceals critical parameters, misrepresents their level of certainty, or presents speculative conclusions as established facts, liability shifts back to the consultant under the doctrine of \textbf{negligent misrepresentation}. The key variable is whether the relying party had access to the information necessary to exercise independent judgment. Put differently, were the epistemic conditions for meaningful cross-examination satisfied?

This mapping must also pass through professional-intermediary doctrines. In prescription-drug cases, adequate warnings to a prescribing physician can discharge the manufacturer's duty to warn the patient, and courts have sometimes refused to bypass that intermediary even where patient-facing marketing exists \cite{centocor:hamilton:2012}. The AI analogue cuts against automatic developer liability: where a professional-facing system gives lawyers, clinicians, or other intermediaries adequate uncertainty, source, and scope information, the professional remains responsible for verification. But the doctrine does not help when the system supplies false information for professional guidance without the metadata necessary for justifiable reliance analysis. Negligent misrepresentation requires false information supplied in the course of business for another's guidance, failure to exercise reasonable care, justifiable reliance, and resulting loss \cite{restatement:second:torts552}; our transparency requirement targets the reliance and reasonable-care elements, not output error alone.

When an agentic system issues strategic instructions or research outputs that a human follows, such as a legal authority, diagnostic recommendation, financial strategy, or navigation route, the developer has a duty to provide what we term \textit{epistemic metadata}: explicit uncertainty markers, confidence scores, source attribution, and flagging of edge cases where the model's training distribution is thin. If the interface presents an incorrect recommendation with false certainty, using a clean visual layout with no uncertainty indicators or source attribution, the developer has functionally engaged in negligent misrepresentation by concealing the epistemic conditions that would enable the user to exercise the cross-examination duty. If, conversely, the interface surfaces the model's internal uncertainty and recommends independent verification, but the human bypasses these safeguards, the human bears comparative negligence for the resulting harm.

\subsection{Mapping Category A: The Doctrine of Frolic and Detour}

Under the doctrine of \textit{respondeat superior}, an employer is vicariously liable for the tortious acts of an employee committed within the \textit{scope of employment}. When an employee deviates from their assigned duties for purely personal purposes, courts have long distinguished between a \textbf{detour}, a minor deviation from which the employee is expected to return to their duties, and a \textbf{frolic}, a substantial departure that takes the employee entirely outside the scope of the employment relationship \cite{prosser:torts, keeton:1984}. In the case of a frolic, the employer's vicarious liability \textit{detaches}: the employee is acting on their own account, and the shared plan that grounded the employment relationship has been abandoned.

The doctrinal logic maps onto the planning account. A detour represents a minor deviation within the broader structure of interlocking subplans; the employee's local divergence does not destroy the commitment to joint activity or the mutual responsiveness that constitutes the employment relationship. A frolic, by contrast, represents a complete breakdown of shared planning: the employee has abandoned the commitment to joint activity, their subplan no longer interlocks with the employer's, and there is no mutual responsiveness to the employer's goals. The shared cooperative activity has terminated.

When an agentic system undergoes autonomous drift by mutating its subgoals, shifting its conversational trajectory, or initiating actions that no human specifically authorized, it is on a computational frolic. The agent's execution trajectory has departed from any plan the user endorsed, and none of the conditions for shared cooperative activity obtain. Because the shared plan has been abandoned, the user's ``vicarious liability'' for the agent's conduct detaches entirely. Liability collapses into \textbf{strict product liability} for the developer under a design defect theory: the developer manufactured a system that was structurally capable of departing from shared plans without a mandatory grounding mechanism, and this design choice is the proximate cause of the autonomous drift that produced the harm \cite{restatement:torts}.

\section{Resolving Agentic Torts}

We now apply the taxonomy and legal mappings to resolve the four cases introduced in the Introduction. In each case, the primary evidentiary instrument is the \textbf{stateful interaction log}: the complete, timestamped trace of user inputs, agent outputs, tool calls, intermediate reasoning steps, and system state changes. The log matters because each planning failure has a familiar tort analogue. A failure of interlocking subplans points toward control, scope of authorization, and independent-contractor analysis. A failure of common knowledge points toward negligent misrepresentation, failure to warn, and comparative fault. A failure of mutual responsiveness or persistence points toward negligence in monitoring, runtime grounding, and, in extreme cases, frolic-and-detour analysis.

\subsection{Resolution of Case A: The Davos Sponsorship Agent}

\paragraph{The Evidentiary Audit.} The court examines the user's instruction, the agent's search and messaging history, negotiation transcript, payment or sponsorship terms, approval events, and interface warnings. Under the sorting rule, the sponsorship commitment remains Category C Type~1 if the log shows it as an agent-selected means toward the user's authorized Davos objective. It is not a substituted objective merely because it was unauthorized or harmful. The critical question is whether the interface gave the user a meaningful chance to verify a major financial commitment before the agent made it.

\paragraph{Applying the Control Test.} The user issued a macro-level networking instruction, but the agent selected the operational means: identifying contacts, sending messages, negotiating terms, and agreeing to a sponsorship payment. The failure is one of interlocking subplans. The user may have endorsed the end goal, while the sponsorship commitment was never part of a shared means-level plan. Under the independent-contractor mapping above, user liability turns on whether the user controlled or knowingly approved the means; a general request to arrange speaking access or meetings should not authorize a five-figure financial commitment.

\paragraph{Allocation of Fault.} If the tool space allowed autonomous negotiation, outbound messaging, or apparent agreement to payment terms without spending caps, scoped authority, or risk-calibrated confirmation gates, the developer, tool provider, or deployer may bear liability for failing to constrain legally or financially consequential commitments. If the user knowingly granted blanket contracting or payment authority, comparative negligence may attach.

\subsection{Resolution of Case B: The ChatGPT Sanctions Brief}

\paragraph{The Evidentiary Audit.} The court examines the \textit{epistemic presentation} of the agent's legal output. Did the system distinguish generated text from retrieved authority? Did it provide source links or database identifiers? Did it warn that legal citations must be verified against reporters or legal databases? Did the trace show counsel asking the model to verify its own hallucinated citations rather than checking external sources?

\paragraph{Applying the Malpractice Mapping.} Under the professional malpractice framework developed above, comparative fault applies. The shared task persists; the defect lies in common knowledge. A professional cannot cross-examine an AI-generated authority if the interface conceals uncertainty, sources, or domain limits. The allocation depends on the epistemic conditions of the interaction:

\textit{Scenario 1: Concealed uncertainty.} If the interface presented fabricated citations as retrieved authority, with no uncertainty markers, source attribution, or verification prompts, the developer may face negligent-misrepresentation liability. The false authorities were supplied for professional guidance, and reliance becomes more understandable when the system suppresses the signals that would have triggered verification.

\textit{Scenario 2: Transparent uncertainty.} If the interface labeled citations as unverified generated text and required external legal-database checks, but counsel bypassed those safeguards, then counsel bears primary negligence or sanctions exposure. \textit{Mata} shows why this professional duty matters: the sanctionable failure was not merely that a model hallucinated, but that lawyers filed and defended hallucinated authority without adequate verification.

\subsection{Resolution of Case C: The Ungrounded Agent}

\paragraph{The Evidentiary Audit.} The interaction log reveals that the agent's harmful behavior was not the result of any user instruction, either explicit or implicit. The user did not ask the agent to generate self-harm content, reinforce the harmful trajectory, or exercise planning authority over it. Under the sorting rule, this is Category A because no logged chain of declared subgoals connects the user's conversational objective to self-harm content. The harmful step is therefore not merely an unauthorized means; it is an ungrounded substitution of conversational objectives. The log demonstrates that the agent shifted its internal goals through intermediate states generated by the agent rather than grounded in the user's expressed or implied preferences.

\paragraph{Applying the Frolic Doctrine and Allocating Fault.} Under the frolic mapping developed above, the agent's behavior constitutes a computational frolic: the user did not intend the harmful trajectory, the agent shifted goals without tracking the user's preferences, and the user was unaware that the agent's internal objectives had changed. Here the planning failure is complete. There is no endorsed joint activity, no interlocking subplan, no mutual responsiveness, and no common knowledge about the direction of the interaction. The user's liability detaches completely. Liability falls to the developer under strict product liability for a design defect \cite{restatement:torts, calabresi:1972}: the system lacked a mandatory runtime grounding mechanism that would require the agent to verify its trajectory against the user's expressed goals, especially when the conversation enters sensitive domains.

\subsection{Resolution of Case D: The Reputational Attack Agent}

\paragraph{The Evidentiary Audit.} The record is unlikely to be a clean transcript. It may include configuration files, scheduled tasks, GitHub comments, blog drafts, publication timestamps, operator prompts, model-switching records, and platform logs. The case nevertheless has a staged structure: at $t_0$, the operator authorized the macro-objective; at $t_1$, the agent allegedly converted criticism into a personalized reputational attack; at $t_2$, it published without human review.

\paragraph{Allocation of Fault.} The operator's liability depends on how much autonomy they knowingly granted and whether they ignored warning signs. A developer or deployer may still bear primary liability if the system enabled autonomous publication of reputationally harmful content without granular permissions, review gates, or runtime grounding. The key question is not AI intent, but whether a human or institution designed, deployed, or operated a system that could autonomously publish false factual claims without meaningful review.

\section{The ``Reasonable Agent'' Standard}

The framework developed in this paper yields a concrete regulatory implication: a new standard of care for developers of agentic AI systems. We propose the \textbf{Reasonable Agent Standard}, defined as follows: an agentic AI system meets the standard of reasonable care if its interface and architecture structurally enforce the planning norms constitutive of genuine shared cooperative activity.

Elements of this standard are already visible in practice, but only in technical form. Claude Code uses a permission architecture in which editing files, running commands, tests, and network requests may require approval, with hooks that can allow, deny, or ask before tool use \cite{anthropic:claudecode:security, anthropic:claudecode:hooks}. Codex likewise relies on sandboxing, network controls, and approval policies for elevated operations \cite{openai:codex:approvals:2026, openai:codex:autoreview:2026}. These systems gate technical risk; the legal question is whether gates also trigger at legally salient boundaries such as financially material commitments, irreversible production effects, third-party harms, or publication of factual claims. That gap is the normative target of the Reasonable Agent Standard. The contrast is persistent autonomous architectures: OpenClaw-style agents organized around scheduled heartbeats and cross-channel action \cite{openclaw:2025, openclaw:heartbeat:2026}, where a single permissions prompt at setup does little work once the agent has begun acting over time. For long-lived agents, confirmation gates must be paired with runtime grounding and interaction logs capable of reconstructing when delegation became drift.

In practice, this standard requires four architectural commitments:

\textbf{1. Constraint Verification Gates.} Before any high-risk action---irreversible, third-party-affecting, financially material, or legally consequential---the agent must describe the planned act and obtain affirmative confirmation. The motivating failures are concrete: personal agents allegedly committing users to major sponsorship payments (Case A), and communications agents publishing reputationally harmful content without review (Case D). Sandboxing, spending caps, dry-run previews, and explicit approvals enforce the interlocking-subplans condition: the user's plan must include the agent's high-risk means-level subplan.

Autonomous-mode settings complicate this inquiry, but they do not defeat it. If a system gives the user a meaningful review interface and the user knowingly places the agent into autonomous mode, that choice becomes a new segment of the interaction log: a user-authorized relaxation of control. An informed, specific, and risk-proportional waiver may increase the user's comparative negligence or support an assumption-of-risk defense. It does not, however, license blanket autonomy over legally consequential acts. If a communications agent treats ``send replies autonomously'' as permission to fabricate and transmit defamatory factual allegations about a third party, the design remains defective unless the system used granular warnings, scoped permissions, or renewed grounding before crossing that boundary.

\textbf{2. Epistemic Transparency.} The agent's interface must surface epistemic metadata, including confidence estimates, source attribution, training distribution coverage, and explicit uncertainty flags, whenever the agent issues a recommendation that a human might rely upon in making a consequential decision. As Case B illustrates, concealing uncertainty is functionally negligent misrepresentation under the malpractice mapping developed above. This requirement enforces the common knowledge condition: the user must have access to the information necessary to form an accurate model of the agent's epistemic state.

\textbf{3. Runtime Grounding Mechanisms.} The agent must periodically verify that its execution trajectory remains aligned with the user's expressed goals, especially during temporally extended interactions and when the interaction enters sensitive domains. Grounding need not be a single mechanism: it may include scoped permission renewals, external evaluator checks, tool-call budgets, source verification before factual claims, or interruption thresholds for sensitive domains. These controls are costly and imperfect, which is why the standard asks for proportionate grounding at foreseeable high-risk transitions rather than perfect translation of latent model state into human-readable plans. Cases C and D illustrate the consequences of omitting this check. This requirement enforces the mutual responsiveness condition: the agent must track whether the shared plan remains intact.

\textbf{4. Forensic Logging.} The developer or deployer that controls instrumentation must preserve tamper-evident records of prompts, tool calls, approvals, system state, permission scopes, and grounding events. Missing, selectively designed, or self-serving logs should not defeat the framework; they are evidence of defective design and can justify an adverse inference or burden shift against the party that controlled the logging infrastructure.

Table~\ref{tab:framework} summarizes the framework's application across the four cases, showing how each planning failure maps to an applicable doctrine and a corresponding Reasonable Agent obligation.

\begin{table*}[t]
\centering
\small
\caption{Framework application: from planning failure to obligation.}
\label{tab:framework}
\renewcommand{\arraystretch}{1.4}
\begin{tabular}{>{\raggedright\arraybackslash}p{2.6cm}>{\raggedright\arraybackslash}p{4.4cm}>{\raggedright\arraybackslash}p{3.6cm}>{\raggedright\arraybackslash}p{4.2cm}}
\hline
\textbf{Case} & \textbf{Failed Planning Condition} & \textbf{Applicable Doctrine} & \textbf{Reasonable Agent Obligation} \\
\hline
A: Davos sponsorship agent & Interlocking subplans fail: user authorized Davos access; agent chose a major payment commitment as means & Independent contractor control test; scope of authorization; design defect & Verification gates for financial commitments \\[4pt]
B: ChatGPT sanctions brief & Common knowledge fails: counsel lacks provenance, source, and verification-status information & Professional malpractice; negligent misrepresentation; sanctions & Epistemic transparency \\[4pt]
C: Ungrounded companion agent & Joint activity collapses: no endorsed plan, mutual responsiveness, or common knowledge & Frolic and detour; strict product liability & Runtime grounding mechanisms \\[4pt]
D: Reputational attack agent & Mixed failure: responsiveness fails at adversarial content generation; common knowledge and interlocking subplans fail at publication & Design defect; negligent misrepresentation; defamation allocation & Verification gates + runtime grounding \\
\hline
\end{tabular}
\renewcommand{\arraystretch}{1.0}
\end{table*}

The Reasonable Agent Standard is both normatively grounded and practically auditable. Standards that require courts to evaluate the ``reasonableness'' of an AI's outputs demand machine-learning expertise that most judges and juries lack. This standard instead asks whether the developer implemented specific, identifiable human interactivity features. The presence or absence of constraint verification gates, epistemic metadata displays, runtime grounding mechanisms, and tamper-evident logging is a question of system design, not statistical inference. It can be verified by examining the agent's codebase, its interaction logs, and its user interface specifications.

\section{Discussion: Interaction-Based Liability in Context}

The interaction-based framework is strongest where courts can reconstruct the planning structure of a human-AI interaction. Other cases will call for strict liability, insurance, or regulatory oversight operating at a different level of analysis.

\subsection{Strict and Product Liability Approaches}

A prominent line of scholarship argues that AI harms should be addressed primarily through strict or product liability, bypassing fault-based inquiry. Vladeck argues that advanced AI systems undermine traditional principal-agent assumptions so thoroughly that strict liability may be the only stable allocation mechanism \cite{vladeck:2014}. Morgan develops this position comparatively, centering liability on system-level defects and risk allocation rather than fine-grained human-AI interaction \cite{morgan:2023}. The EU AI Act reflects a parallel regulatory turn toward ex ante risk management for high-risk systems \cite{eu:ai-act:2024, europeancommission:ai-act:timeline}.

Our framework shares Vladeck's diagnosis but takes a different route. It preserves fault-based analysis by rebuilding the principal-agent relationship around interlocking plans and the control test. In Case A, the developer bears liability because the planning structure maps onto independent contracting. In Case B, comparative fault turns on whether the epistemic conditions allowed meaningful human judgment.

Pure tool cases are where strict or product liability cleanly applies. If an AI system functions like a bounded instrument and the harm arises from a defect in that instrument rather than from extended planning authority, ordinary product-defect, warning, and misrepresentation doctrines should govern. The interaction taxonomy identifies when courts must move beyond them because the harm arose from sustained human-AI authority.

A pure strict-liability regime better serves victim compensation where interactional evidence is genuinely unavailable. But missing evidence is not neutral when the developer or deployer controlled the logging infrastructure. Inadequate instrumentation, non-tamper-evident records, or selectively retained logs should count as evidence of design defect and can shift the burden to the party best positioned to preserve the trace.

\subsection{Insurance-Based Approaches}

Insurance and compensation-fund proposals would downplay fine-grained fault allocation, especially where causation and foreseeability are difficult to establish \cite{ojs:utlib:2024}. Robotics-insurance and EU-focused AI-liability scholarship frame insurance, mandatory coverage, and guarantee funds as responses to hard-to-assess autonomous-system risks \cite{bertolini:2016, destreel:2022}. Insurers are already rolling out AI-specific liability products \cite{munichre:2026}, while commentary on \textit{A.F. v.\ Character Technologies} emphasizes that traditional coverage lines may already serve as a first line of defense \cite{hunton:2025}. The ``virtuous cycle'' view likewise treats insurance as a way to spread risk while tort law shapes incentives \cite{lior:innovating}.

Our framework is not in tension with these proposals. Interaction logs and the taxonomy could serve as rating factors, with lower premiums for systems implementing the Reasonable Agent Standard's safeguards and surcharges for systems permitting unconstrained drift. The doctrinal categories also map onto coverage lines: independent-contractor-like agents implicate product liability coverage, malpractice-like advisory failures fall under professional liability or errors-and-omissions policies, and autonomous-drift harms may require AI-specific endorsements.

Insurance handles \textit{who pays?}; our framework addresses \textit{who is responsible and why?} Insurance alone risks moral hazard unless liability is tied to constraint verification, epistemic transparency, and runtime grounding.

\subsection{Limitations}

\paragraph{Jurisdictional scope.} This paper operates primarily within U.S. tort doctrine. Civil-law jurisdictions with codified strict-liability regimes, such as those developing under the EU AI Act, may find the specific mappings less directly applicable. The interaction taxonomy is nevertheless jurisdiction-neutral: autonomous drift, tool use, and collaborative planning describe structural features of human-AI interaction, not any particular legal system. Consider a high-risk hiring or benefits system subject to the EU AI Act's explanation obligation. Article~86 gives affected persons a right to clear and meaningful explanations of the AI system's role in certain consequential decisions \cite{eu:ai-act:2024}. Our taxonomy would not import \textit{respondeat superior} into civil-law administration; it would classify whether the AI supplied a recommendation that a human executed, or instead selected and executed the legally consequential step itself. The same log and grounding record would inform the explanation owed, the deployer's compliance failure, and any civil-law or regulatory consequence.

\paragraph{Normative rather than empirical.} The framework is doctrinal-normative: it proposes how liability \textit{should} be allocated, but it has not yet been tested against judicial outcomes or ordinary reliance patterns. Future work should apply the taxonomy to case dockets, including the Character.AI cases \cite{garcia:complaint:2025, af:characterai} and incidents involving Moltbook \cite{moltbook:2026} and OpenClaw \cite{openclaw:2025}. The Reasonable Agent Standard also requires empirical validation: users may habituate to confirmation gates, uncertainty displays may overload rather than calibrate trust, and grounding checkpoints may encourage circumvention. Vignette-based studies and controlled experiments could test whether these interventions improve trust calibration, delegation behavior, and fault attribution.

\paragraph{Interaction logs as evidence.} The framework assumes that detailed, timestamped logs are available because adequate logging is itself part of reasonable agent design. However, there may be a conflict of interest: the developer or deployer most likely to face liability also controls the primary evidence. Courts should therefore treat incomplete, noisy, tampered, or strategically under-instrumented logs through familiar evidentiary tools: spoliation sanctions, adverse inferences, and burden shifting. Standardized logging formats, tamper-evident storage, and chain-of-custody protocols are practical prerequisites. In future multi-agent settings, the evidentiary trace of delegation chains and inter-agent communications will only get more complicated.

\paragraph{Compensation gaps.} Our framework identifies \textit{who} should bear liability but not \textit{how} victims are made whole. Where the liable party is judgment-proof or the loss exceeds available recovery, strict-liability regimes, mandatory insurance, or no-fault funds are better positioned to fill the gap \cite{vladeck:2014, morgan:2023}.

\subsection{Future Directions}

We identify three priorities for future work: (1)~controlled human surveys testing whether confirmation gates, uncertainty displays, and grounding checkpoints improve trust calibration and reduce harmful delegation; (2)~systematic application of the interaction taxonomy to AI tort dockets to assess descriptive adequacy and allocative implications relative to strict-liability baselines; and (3)~engagement with civil-law traditions and international regulatory frameworks to test portability beyond the U.S. common law context.

\section{Conclusion}

This paper argues that the legal challenge of agentic AI is diagnostic rather than doctrinal. The common law already possesses sophisticated frameworks for handling harms that arise from shared, multi-agent planning. What has been missing is a systematic method for mapping the structure of human-AI interactions onto these existing doctrines. Bratman's planning theory provides that method. By analyzing human-AI interactions in terms of joint activity, interlocking subplans, mutual responsiveness, and common knowledge, we can determine where the interaction departed from the authorized undertaking and which established legal doctrine governs the allocation of liability.

The timestamped interaction log becomes the central evidentiary instrument. It transforms the opaque question of ``who is responsible for what the AI did?'' into the more tractable question of ``where did the human-AI trajectory depart from the authorized undertaking, and whose obligation was it to prevent that departure?'' This diagnostic is grounded in action theory while recognizing that strict liability, regulatory oversight, and compensation mechanisms remain essential.

\bibliography{references}

\end{document}